\documentclass[conference]{IEEEtran}
\IEEEoverridecommandlockouts

\usepackage{amsmath,amssymb,amsfonts}
\usepackage{graphicx}
\usepackage{booktabs}
\usepackage{balance}
\usepackage{textcomp}
\usepackage{cite}
\usepackage{xcolor}
\definecolor{refblue}{RGB}{0,0,190}

\usepackage{hyperref}
\hypersetup{
  colorlinks = true,
  linkcolor  = blue,
  citecolor  = blue,
  urlcolor   = blue,
  filecolor  = blue,
}
\usepackage{tikz}
\usepackage{enumitem}
\newcommand{\R}{\mathbb{R}}
\newcommand{\SO}{\mathrm{SO}}

\newcommand{\tr}{\operatorname{tr}}
\newcommand{\rank}{\operatorname{rank}}

\AtBeginDocument{%
  \setlength{\abovedisplayskip}{2pt plus 1pt minus 1pt}%
  \setlength{\belowdisplayskip}{2pt plus 1pt minus 1pt}%
  \setlength{\abovedisplayshortskip}{0pt plus 1pt}%
  \setlength{\belowdisplayshortskip}{1pt plus 1pt}%
}
\setlist{topsep=2pt,itemsep=1pt,parsep=0pt}

\newtheorem{problem}{Problem}

\newcommand{\Cay}{\mathrm{Cay}}
\newcommand{\Exp}{\mathrm{Exp}}
\newcommand{\Ret}{\mathcal{R}}
\newcommand{\bvec}{\mathbf{b}}

\newcommand{\wed}{{}^{\wedge}}

\newcommand{\W}{{\mathrm{W}}}

\AtBeginDocument{\let\oldthebibliography\thebibliography
  \renewcommand{\thebibliography}[1]{\oldthebibliography{#1}\small}}

\begin{document}

\title{A QCQP-Representable IMU Pre-Integration Factor\\for Certifiable State Estimation}

\author{\IEEEauthorblockN{Utkarsh Rai, Zhexin Xu, Bang-Shien Chen, and David Rosen}%
\thanks{The authors are with the Robust Autonomy Lab, Institute for Experiential
Robotics, Northeastern University, 360 Huntington Ave, Boston, MA 02115, USA.}}

\maketitle

\begin{abstract}
We propose a QCQP-representable IMU pre-integration factor that enables
certifiable estimation with pre-integrated inertial measurements. To the best
of our knowledge, this is the first work to directly incorporate IMU
pre-integration into certifiable estimation. 
Inertial sensing is a common and reliable modality in robotics, and incorporating it broadens the practical scope of certifiable estimation.
The main challenges are obtaining the required algebraic structure and a
sufficiently tight convex relaxation. Standard IMU pre-integration relies on
the exponential map, which does not admit an exact polynomial representation.
Moreover, obtaining a QCQP formulation requires auxiliary lifting variables,
for which the standard semidefinite programming (SDP) relaxation can be loose.
We address these issues by deriving an IMU pre-integration factor based on the
Cayley map and an explicit set of redundant constraints that tighten the
resulting relaxation.
To validate the proposed factor, we apply it to certifiable GNSS--IMU
smoothing and evaluate it on synthetic and real-world data. The results show
that the proposed formulation yields tight relaxations and solves the resulting
estimation problems to verified global optimality.
\end{abstract}

\section{Introduction}
\label{sec:intro}

State estimation in robotics is conventionally formulated as an optimization problem. Local
optimization methods such as Gauss--Newton are widely used for their efficiency and ease of
use. However, the resulting optimization problems are generally nonconvex
(e.g., due to rotation constraints), so local methods can converge to local minima and may fail without warning.

An alternative is provided by certifiable estimation methods, which can recover
a verified globally optimal solution without relying on an initialization.
The key machinery behind these methods consists of formulating the estimation problem as a
quadratically constrained quadratic program (QCQP)
and apply a semidefinite relaxation, yielding a convex
semidefinite program (SDP). For a minimization problem, the optimal value of the
SDP provides a global lower bound on that of the original QCQP. When the optimal
solution matrix $\hat{Z}$ of the SDP has rank one, i.e.,
$\rank(\hat{Z})=1$, the relaxation is \emph{tight}. Its rank-one factorization
$\hat{Z}=\hat{z}\hat{z}^{\top}$ then recovers a globally optimal solution
$\hat{z}$ of the original QCQP.

Certifiable methods have been widely applied to robotics state estimation problems, including
pose graph optimization~\cite{sesync}, rotation averaging~\cite{hartley,shonan}, outlier-robust
estimation~\cite{yangcarlone}, range-aided SLAM~\cite{cora}, and pose estimation using the
Cayley map~\cite{barfoot}. Together, these applications span a broad range of sensing
modalities. Certifiable estimation has also been generalized through the framework of
certifiable factor graphs~\cite{cfgo}. Despite this breadth, measurements from inertial
measurement units (IMUs), among the most common and reliable sensors used in autonomous
navigation, have not yet been incorporated into certifiable estimation frameworks.

\begin{figure}[t]
\centering
\includegraphics[width=\columnwidth]{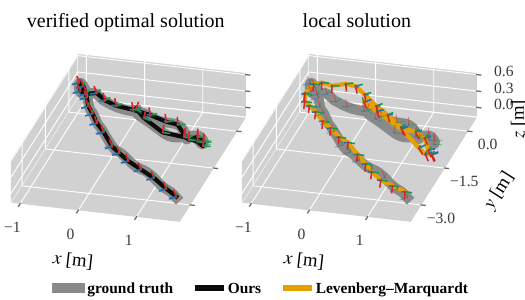}
\caption{
\textbf{Local and certifiable solutions for GNSS--IMU smoothing.}
Accurate GNSS position measurements keep the poorly initialized local solution
(orange) relatively close to the ground truth in position, but its orientation
is gravity-flipped. The certifiable estimator (black) requires no trajectory
initialization and returns a verified globally optimal solution that closely
matches the ground truth in both position and orientation.
}
\label{fig:teaser}
\end{figure}

The main obstacle to formulating a certifiable estimator with inertial measurements is
obtaining the required QCQP structure. Preintegration is the standard approach to incorporating
high-rate inertial data into estimation frameworks. We therefore consider certifiable
estimation with inertial preintegration. Conventional preintegration updates rotations using
the exponential map, which does not admit an exact polynomial representation. In contrast, the
Cayley map has been shown to yield a natural QCQP structure but has not been considered for
IMU preintegration.

Even after a QCQP formulation is obtained, its semidefinite relaxation need not be tight. This
issue is particularly relevant to inertial preintegration because its QCQP formulation requires
auxiliary lifting variables, which may make redundant constraints necessary for obtaining a
tight relaxation. 
Such constraints have been derived for relative pose
estimation~\cite{briales}, extrinsic calibration~\cite{wise2026certifiably}, and certifiable
trajectory estimation using the Cayley map~\cite{barfoot}, but not for the
state--bias coupling liftings required by
inertial preintegration.

In this work, we address these challenges by introducing the first method to
\emph{directly}\footnote{
By \emph{directly}, we mean using IMU pre-integration itself in the estimator,
rather than first converting IMU data into relative pose measurements and
then using them as pose-to-pose factors. Existing certifiable pose-estimation
methods can already use IMU information this way, but this loses the
\emph{tightly coupled} inertial model, including velocity and bias estimation,
which are central to inertial navigation using IMUs.
}
incorporate inertial measurements into certifiable estimation.

\paragraph{Contributions}
\begin{itemize}[leftmargin=1.6em,itemsep=1pt,topsep=2pt]

\item \textbf{QCQP-Representable IMU Preintegration Factor:}
We derive a Cayley-based IMU preintegration factor that admits a QCQP
representation.

\item \textbf{Tightening the SDP Relaxation:} 
We derive redundant constraints for the rotation, wedge, and vector liftings that strengthen the SDP relaxation of the proposed QCQP formulation.

\item \textbf{Validations:} 
We evaluate the derived redundant constraints through synthetic ablation studies of SDP relaxation exactness under different configurations. We also apply the factor to certifiable GNSS--IMU smoothing on real-world data and compare its performance with local optimization.

\end{itemize}

\paragraph{Organization}
Sec.~\ref{sec:prelim} reviews retractions on $\SO(3)$ and the QCQP--SDP
certification framework. Sec.~\ref{sec:expreint} introduces standard IMU
pre-integration, the resulting inertial MLE, and the obstruction to a QCQP
representation. Secs.~\ref{sec:preint} and~\ref{sec:redundant} present our
main technical contributions: the Cayley-based pre-integration factor and the
set of redundant constraints. Sec.~\ref{sec:problem} applies
the proposed factor to GNSS--IMU smoothing, and Sec.~\ref{sec:exp} presents
the experimental evaluation.

\section{Preliminaries}
\label{sec:prelim}

This section reviews the two main ingredients used in the rest of the
paper. We first introduce the exponential and Cayley retractions on
$\SO(3)$, focusing on the properties needed for IMU pre-integration.
We then review QCQP-based certifiable estimation and its semidefinite
relaxation, which provide the optimization framework used later.

\subsection{Retractions on the $\SO(3)$ Manifold}
\label{sec:cayley}

A retraction $\Ret:\R^3\to\SO(3)$ maps a tangent vector to a rotation and satisfies
$\Ret(\mathbf{0})=\mathbf{I}$. Its right Jacobian $\mathbf{J}_{\Ret}$ relates an additive
perturbation $\delta\boldsymbol{\phi}$ in the tangent space to a multiplicative perturbation on
the manifold. We use two properties of these retractions:
\begin{equation}
\begin{aligned}
\Ret(\boldsymbol{\phi}+\delta\boldsymbol{\phi})
&= \Ret(\boldsymbol{\phi})\,
   \Ret\!\left( \mathbf{J}_{\Ret}(\boldsymbol{\phi})\,
   \delta\boldsymbol{\phi} \right) + \mathcal{O}(\|\delta\boldsymbol{\phi}\|^2), \\
\Ret(\mathbf{R}^\top \boldsymbol{\phi})
&= \mathbf{R}^\top \Ret(\boldsymbol{\phi}) \mathbf{R},
   \qquad \mathbf{R}\in\SO(3),
\end{aligned}
\label{eq:retprops}
\end{equation}
which are the first-order composition rule and the adjoint identity, respectively.

The exponential map $\Exp(\boldsymbol{\phi})=\exp(\boldsymbol{\phi}\wed)$ associates a tangent
vector with a rotation and coincides with the standard matrix exponential (Rodrigues' formula):
\begin{equation}
\Exp(\boldsymbol{\phi})
=
\mathbf{I}
+
\frac{\sin\|\boldsymbol{\phi}\|}{\|\boldsymbol{\phi}\|}\boldsymbol{\phi}\wed
+
\frac{1-\cos\|\boldsymbol{\phi}\|}{\|\boldsymbol{\phi}\|^2}(\boldsymbol{\phi}\wed)^2 .
\label{eq:rodrigues}
\end{equation}
Its right Jacobian $\mathbf{J}_r(\boldsymbol{\phi})$ is likewise a function of
$\sin\|\boldsymbol{\phi}\|$ and $\cos\|\boldsymbol{\phi}\|$~\cite{forster}. These trigonometric
terms make $\Exp$ non-polynomial in $\boldsymbol{\phi}$ and prevent a direct QCQP
representation.

The Cayley map is an alternative retraction. In the half-angle convention it reads
\begin{equation}
\Cay(\boldsymbol{\phi})
=
\left( \mathbf{I} - \tfrac{1}{2}\boldsymbol{\phi}\wed \right)^{-1}
\left( \mathbf{I} + \tfrac{1}{2}\boldsymbol{\phi}\wed \right),
\qquad
\boldsymbol{\phi} = 2\tan(\theta/2)\,\mathbf{a},
\label{eq:cayley}
\end{equation}
where $\boldsymbol{\phi}$ maps to a rotation of angle $\theta$ about the unit axis $\mathbf{a}$,
and its right Jacobian is
\begin{equation}
\mathbf{J}_{\Cay}(\boldsymbol{\phi})
=
\frac{\mathbf{I} - \tfrac{1}{2}\boldsymbol{\phi}\wed}
     {1 + \tfrac{1}{4}\|\boldsymbol{\phi}\|^2}.
\label{eq:cayjac}
\end{equation}
The Cayley map satisfies $\Cay(\boldsymbol{\phi})\in\SO(3)$ for every
$\boldsymbol{\phi}\in\R^3$, and agrees with the exponential map to second order,
$\Cay(\boldsymbol{\phi})=\Exp(\boldsymbol{\phi})+\mathcal{O}(\|\boldsymbol{\phi}\|^3)$. Unlike
$\Exp$, the Cayley map is rational in $\boldsymbol{\phi}$, so its denominator can be cleared
algebraically.

\subsection{Certifiable Estimation}
\label{sec:certifiable_estimation}

Many maximum-likelihood estimation problems in robotics can be formulated as
quadratically constrained quadratic programs (QCQPs)~\cite{boyd,shor}:
\begin{equation}
\label{eq:qcqp}
\begin{aligned}
  f_{\mathrm{QCQP}}^{\star}
  =
  \min_{\mathbf{x} \in \mathbb{R}^{n}} \quad
    & \mathbf{x}^{\top} \mathbf{Q}\mathbf{x} \\
  \mathrm{s.\,t.} \quad
    & \mathbf{x}^{\top} \mathbf{A}_j \mathbf{x} = b_j,
      \quad j = 1, \dots, m,
\end{aligned}
\end{equation}
where $\mathbf{x} \in \mathbb{R}^{n}$ is the optimization variable,
$\mathbf{Q}$ is a symmetric cost matrix, and each pair
$(\mathbf{A}_j,b_j)$, with $\mathbf{A}_j$ symmetric, defines a quadratic
equality constraint. In robotic estimation, these constraints often define
a nonconvex feasible set (e.g., rotation constraints). A local solver applied
to~\eqref{eq:qcqp} can therefore guarantee only local optimality, and its
solution may depend on the initialization.

To obtain a convex relaxation, we introduce the lifted matrix
$\mathbf{Z}=\mathbf{x}\mathbf{x}^{\top}$. The quadratic terms then become
linear in $\mathbf{Z}$, while the relation
$\mathbf{Z}=\mathbf{x}\mathbf{x}^{\top}$ is represented by a rank-one
constraint. Dropping this constraint and allowing
$\mathbf{Z}\in\mathbb{S}^{n}_{+}$ gives the semidefinite program (SDP):
\begin{equation}
\label{eq:sdp}
\begin{aligned}
  f_{\mathrm{SDP}}^{\star}
  =
  \min_{\mathbf{X} \in \mathbb{S}^{n}_{+}} \quad
    & \operatorname{tr}(\mathbf{Q}\mathbf{Z}) \\
  \mathrm{s.\,t.} \quad
    & \operatorname{tr}(\mathbf{A}_j\mathbf{Z}) = b_j,
      \quad j = 1, \dots, m,
\end{aligned}
\end{equation}
known as Shor's relaxation~\cite{shor}, where
$\mathbb{S}^{n}_{+}$ denotes the set of $n\times n$ symmetric positive
semidefinite matrices. Since~\eqref{eq:sdp} is a relaxation
of~\eqref{eq:qcqp}, we have
\begin{equation}
\label{eq:bound}
  f_{\mathrm{SDP}}^{\star} \leq f_{\mathrm{QCQP}}^{\star}.
\end{equation}
The SDP is convex and can be solved to global optimality using standard
convex optimization methods. When~\eqref{eq:bound} holds with equality,
the relaxation is said to be \emph{tight}.

Suppose a minimizer $\mathbf{Z}^{\star}$ of~\eqref{eq:sdp} has rank one.
It then admits the factorization
\[
  \mathbf{Z}^{\star} = \mathbf{x}^{\star}\mathbf{x}^{\star\top}.
\]
The vector $\mathbf{x}^{\star}$ is feasible for~\eqref{eq:qcqp}, and
\[
  \mathbf{x}^{\star\top}\mathbf{Q}\mathbf{x}^{\star}
  = \operatorname{tr}(\mathbf{Q}\mathbf{Z}^{\star})
  = f_{\mathrm{SDP}}^{\star}.
\]
With~\eqref{eq:bound}, this gives
$f_{\mathrm{QCQP}}^{\star}=f_{\mathrm{SDP}}^{\star}$, so
$\mathbf{x}^{\star}$ is a global minimizer of~\eqref{eq:qcqp}. Few
relaxations in robotic estimation are tight by construction~\cite{sesync},
so in practice the relaxation may need to be
\emph{tightened}~\cite{autotight} before a rank-one solution can be expected.


\section{Exponential-Map IMU Pre-Integration}
\label{sec:expreint}

IMU pre-integration summarizes high-rate inertial measurements as a single
factor between consecutive keyframes. Standard exponential-map
pre-integration, however, cannot be written directly in QCQP form. This section
reviews standard IMU pre-integration~\cite{forster}, defines the corresponding
factor, and identifies the nonpolynomial terms.

\subsection{Pre-Integration Model}
\label{sec:preintegration}

An IMU measures angular velocity
$\tilde{\boldsymbol{\omega}}(t)\in\mathbb R^3$ and specific force
$\tilde{\mathbf a}(t)\in\mathbb R^3$, both expressed in the IMU frame. The
measurement model is
\begin{subequations}
\label{eq:imu_measurement_model}
\begin{align}
\tilde{\boldsymbol{\omega}}(t)
&=
\boldsymbol{\omega}(t)
+\mathbf b_g(t)
+\boldsymbol{\eta}^g(t), \\
\tilde{\mathbf a}(t)
&=
\mathbf R^\top(t)
\bigl(\mathbf a(t)-\mathbf g^{\mathrm W}\bigr)
+\mathbf b_a(t)
+\boldsymbol{\eta}^a(t).
\end{align}
\end{subequations}
Here, $\mathbf R(t)\in\SO(3)$ maps vectors from the IMU frame to the world
frame, $\boldsymbol{\omega}(t)$ is the angular velocity expressed in the IMU
frame, $\mathbf a(t)$ is the acceleration expressed in the world frame, and
$\mathbf g^{\mathrm W}$ is the gravity vector. The vectors
$\mathbf b_g(t)$ and $\mathbf b_a(t)$ are the gyroscope and accelerometer
biases. The measurement noises $\boldsymbol{\eta}^g(t)$ and
$\boldsymbol{\eta}^a(t)$ are modeled as zero-mean Gaussian white noise. The
biases follow the random-walk models
\[
\dot{\mathbf b}_g(t)=\boldsymbol{\eta}^{bg}(t),
\qquad
\dot{\mathbf b}_a(t)=\boldsymbol{\eta}^{ba}(t),
\]
where $\boldsymbol{\eta}^{bg}(t)$ and
$\boldsymbol{\eta}^{ba}(t)$ are the corresponding driving noises.

Let $t_k$ denote the IMU sampling times, with
$\tilde{\boldsymbol{\omega}}_k
\triangleq\tilde{\boldsymbol{\omega}}(t_k)$ and
$\tilde{\mathbf a}_k\triangleq\tilde{\mathbf a}(t_k)$.
We assume a constant sampling interval
$\Delta t\triangleq t_{k+1}-t_k$, so that
$\Delta t_{ij}\triangleq t_j-t_i=(j-i)\Delta t$.
The biases are treated as constant between keyframes $i$ and $j$.

Pre-integration is performed at the nominal bias
\[
\bar{\mathbf b}
\triangleq
\begin{bmatrix}
\bar{\mathbf b}_g^\top &
\bar{\mathbf b}_a^\top
\end{bmatrix}^{\!\top},
\]
where $\bar{\mathbf b}_g$ and $\bar{\mathbf b}_a$ are the nominal gyroscope
and accelerometer biases. The pre-integrated measurements are
\begin{subequations}
\label{eq:preintegrated_deltas}
\begin{align}
\Delta\tilde{\mathbf R}_{ij}(\bar{\mathbf b})
&\triangleq
\prod_{k=i}^{j-1}
\Exp\!\left(
(\tilde{\boldsymbol{\omega}}_k-\bar{\mathbf b}_g)\Delta t
\right), \\
\Delta\tilde{\mathbf v}_{ij}(\bar{\mathbf b})
&\triangleq
\sum_{k=i}^{j-1}
\Delta\tilde{\mathbf R}_{ik}(\bar{\mathbf b})
(\tilde{\mathbf a}_k-\bar{\mathbf b}_a)\Delta t, \\
\Delta\tilde{\mathbf p}_{ij}(\bar{\mathbf b})
&\triangleq
\sum_{k=i}^{j-1}
\left(
\Delta\tilde{\mathbf v}_{ik}(\bar{\mathbf b})\Delta t
+\frac{1}{2}
\Delta\tilde{\mathbf R}_{ik}(\bar{\mathbf b})
(\tilde{\mathbf a}_k-\bar{\mathbf b}_a)\Delta t^2
\right).
\end{align}
\end{subequations}
The partial increments are initialized as
$\Delta\tilde{\mathbf R}_{ii}(\bar{\mathbf b})=\mathbf I$,
$\Delta\tilde{\mathbf v}_{ii}(\bar{\mathbf b})=\mathbf 0$, and
$\Delta\tilde{\mathbf p}_{ii}(\bar{\mathbf b})=\mathbf 0$.

The quantities in~\eqref{eq:preintegrated_deltas} are computed from noisy IMU
measurements. For use in the estimator, the pre-integrated measurement is
separated into the relative motion predicted by the states and the noise
accumulated over the integration interval. Let
$\mathbf R_\ell\in\SO(3)$ and
$\mathbf v_\ell,\mathbf p_\ell\in\mathbb R^3$
denote the orientation, velocity, and position at keyframe
$\ell\in\{i,j\}$. The measurement model is
\begin{subequations}
\label{eq:preintegration_noise_model}
\begin{align}
\Delta\tilde{\mathbf R}_{ij}(\bar{\mathbf b})
&=
\mathbf R_i^\top\mathbf R_j
\Exp(\delta\boldsymbol\phi_{ij}), \\
\Delta\tilde{\mathbf v}_{ij}(\bar{\mathbf b})
&=
\mathbf R_i^\top
\bigl(
\mathbf v_j-\mathbf v_i-\mathbf g^{\mathrm W}\Delta t_{ij}
\bigr)
+\delta\mathbf v_{ij}, \\
\Delta\tilde{\mathbf p}_{ij}(\bar{\mathbf b})
&=
\mathbf R_i^\top
\bigl(
\mathbf p_j-\mathbf p_i-\mathbf v_i\Delta t_{ij}
-\tfrac12\mathbf g^{\mathrm W}\Delta t_{ij}^2
\bigr)
+\delta\mathbf p_{ij}.
\end{align}
\end{subequations}
The rotation error $\delta\boldsymbol\phi_{ij}\in\mathbb R^3$ enters
multiplicatively, while the velocity and position errors
$\delta\mathbf v_{ij},\delta\mathbf p_{ij}\in\mathbb R^3$ enter additively.
Their stacked form is
\[
\boldsymbol\delta_{ij}
\triangleq
\begin{bmatrix}
\delta\boldsymbol\phi_{ij}^\top &
\delta\mathbf v_{ij}^\top &
\delta\mathbf p_{ij}^\top
\end{bmatrix}^{\!\top}
\sim
\mathcal N
\bigl(
\mathbf 0,\boldsymbol\Sigma^{\mathrm f}_{ij}
\bigr),
\]
where
$\boldsymbol\Sigma^{\mathrm f}_{ij}\in\mathbb R^{9\times9}$
is the covariance obtained by propagating the IMU measurement noise through
the pre-integration. Its off-diagonal blocks capture correlations among the
three errors. See~\cite{forster} for the full derivation of the error model
and covariance propagation.

Although the measurements in~\eqref{eq:preintegrated_deltas} are computed at
$\bar{\mathbf b}$, the bias remains an estimation variable. Let
\[
\mathbf b
\triangleq
\begin{bmatrix}
\mathbf b_g^\top &
\mathbf b_a^\top
\end{bmatrix}^{\!\top},
\qquad
\delta\mathbf b
\triangleq
\mathbf b-\bar{\mathbf b}
=
\begin{bmatrix}
\delta\mathbf b_g^\top &
\delta\mathbf b_a^\top
\end{bmatrix}^{\!\top}.
\]
Recomputing the pre-integration for every value of $\mathbf b$ would be
expensive, so the change is approximated to first order.

The matrices
$\mathbf J^R_{ij}$,
$\mathbf J^{vg}_{ij}$,
$\mathbf J^{va}_{ij}$,
$\mathbf J^{pg}_{ij}$, and
$\mathbf J^{pa}_{ij}\in\mathbb R^{3\times3}$
are the bias Jacobians evaluated at $\bar{\mathbf b}$. The superscripts denote
the pre-integrated quantity and the corresponding bias. For example,
$\mathbf J^{vg}_{ij}$ is the Jacobian of the pre-integrated velocity with
respect to the gyroscope bias. The rotation Jacobian
$\mathbf J^R_{ij}$ maps the gyroscope-bias perturbation to the tangent-space
correction
\[
\boldsymbol\xi_{ij}
\triangleq
\mathbf J^R_{ij}\delta\mathbf b_g.
\]
The bias-corrected measurements are
\begin{subequations}
\label{eq:bias_update}
\begin{align}
\Delta\tilde{\mathbf R}_{ij}(\mathbf b)
&\approx
\Delta\tilde{\mathbf R}_{ij}(\bar{\mathbf b})
\Exp(\boldsymbol\xi_{ij}), \\
\Delta\tilde{\mathbf v}_{ij}(\mathbf b)
&\approx
\Delta\tilde{\mathbf v}_{ij}(\bar{\mathbf b})
+\mathbf J^{vg}_{ij}\delta\mathbf b_g
+\mathbf J^{va}_{ij}\delta\mathbf b_a, \\
\Delta\tilde{\mathbf p}_{ij}(\mathbf b)
&\approx
\Delta\tilde{\mathbf p}_{ij}(\bar{\mathbf b})
+\mathbf J^{pg}_{ij}\delta\mathbf b_g
+\mathbf J^{pa}_{ij}\delta\mathbf b_a.
\end{align}
\end{subequations}
These Jacobians are computed recursively during pre-integration.

\subsection{Pre-Integration Factor}
\label{sec:expresid}

The pre-integration factor compares the relative motion predicted by the
states with the bias-corrected measurements in~\eqref{eq:bias_update}. The
rotation residual uses the geodesic error on $\SO(3)$, while the velocity and
position residuals use Euclidean differences:
\begin{subequations}
\label{eq:exp_residuals}
\begin{align}
\mathbf r^\phi_{ij}
&=
\mathrm{Log}\!\left(
\Delta\tilde{\mathbf R}_{ij}(\mathbf b)^\top
\mathbf R_i^\top\mathbf R_j
\right),
\label{eq:resphi}\\
\mathbf r^v_{ij}
&=
\mathbf R_i^\top
\bigl(
\mathbf v_j-\mathbf v_i-\mathbf g^{\mathrm W}\Delta t_{ij}
\bigr)
-\Delta\tilde{\mathbf v}_{ij}(\mathbf b),
\label{eq:resv}\\
\mathbf r^p_{ij}
&=
\mathbf R_i^\top
\bigl(
\mathbf p_j-\mathbf p_i-\mathbf v_i\Delta t_{ij}
-\tfrac12\mathbf g^{\mathrm W}\Delta t_{ij}^2
\bigr)
-\Delta\tilde{\mathbf p}_{ij}(\mathbf b).
\label{eq:resp}
\end{align}
\end{subequations}

Stacking the residuals gives
\[
\mathbf r_{ij}
\triangleq
\begin{bmatrix}
(\mathbf r^\phi_{ij})^\top &
(\mathbf r^v_{ij})^\top &
(\mathbf r^p_{ij})^\top
\end{bmatrix}^{\!\top}
\in\mathbb R^9.
\]
Under the Gaussian noise model in~\eqref{eq:preintegration_noise_model}, the
pre-integration factor is
\begin{equation}
F^{\mathrm{Exp}}_{ij}
\triangleq
\frac{1}{2}
\left\|\mathbf r_{ij}\right\|^2_
{\left(\boldsymbol\Sigma^{\mathrm f}_{ij}\right)^{-1}},
\label{eq:expfactor}
\end{equation}
where
$\|\mathbf r_{ij}\|^2_{(\boldsymbol\Sigma^{\mathrm f}_{ij})^{-1}}$
is the squared Mahalanobis norm and
$(\boldsymbol\Sigma^{\mathrm f}_{ij})^{-1}$ is the pre-integration
information matrix.

\subsection{Technical Gap in Certifiable Estimation}
\label{sec:pipeline}

The rotation residual $\mathbf r^\phi_{ij}$ in~\eqref{eq:resphi} cannot be
written exactly as a polynomial for two reasons. First, the geodesic residual
contains the logarithmic map, which is nonpolynomial. We therefore replace it
with the chordal residual
\begin{equation}
\left\|
\mathbf R_j
-
\mathbf R_i
\Delta\tilde{\mathbf R}_{ij}(\bar{\mathbf b})
\Exp(\boldsymbol\xi_{ij})
\right\|_F^2.
\label{eq:expobstruction}
\end{equation}

The second issue is the exponential map used for bias correction. By
Rodrigues' formula~\eqref{eq:rodrigues},
$\Exp(\boldsymbol\xi_{ij})$ contains trigonometric functions of
$\|\boldsymbol\xi_{ij}\|$. Since
$\boldsymbol\xi_{ij}=\mathbf J^R_{ij}\delta\mathbf b_g$ depends on the unknown
gyroscope bias, the chordal residual remains nonpolynomial. Therefore, the
standard exponential-map pre-integration factor is not directly
QCQP-representable. The next section replaces this exponential-map correction
with the Cayley map.

\section{A QCQP-Representable Pre-Integration Factor}
\label{sec:preint}

This section develops the Cayley-based IMU pre-integration factor used in our
QCQP formulation. We first derive a polynomial rotation residual. We then
introduce auxiliary variables for the remaining state--bias products and
show the resulting quadratic cost and constraints.

\subsection{Cayley Pre-Integrated Measurements}
\label{sec:cayint}

The Cayley map is a rational alternative to the exponential map. Because it
has analogous first-order composition and conjugation properties, the
pre-integration derivation in Sec.~\ref{sec:preintegration} carries over after
replacing $\Exp$ and its right Jacobian with $\Cay$ and the Cayley right
Jacobian in~\eqref{eq:cayjac}. We denote the resulting nominal pre-integrated
measurements by $\Delta\bar{\mathbf R}_{ij}$,
$\Delta\bar{\mathbf v}_{ij}$, and $\Delta\bar{\mathbf p}_{ij}$.

Let $\mathbf J^R_{ij}\in\mathbb R^{3\times3}$ denote the accumulated Jacobian
of the Cayley pre-integrated rotation with respect to the gyroscope bias.
Under the first-order noise and bias model, the relative rotation satisfies
\begin{equation}
\mathbf{R}_i^\top \mathbf{R}_j
=
\Delta\bar{\mathbf{R}}_{ij}\,
\Cay(\boldsymbol{\xi}_{ij})\,
\Cay(-\delta\boldsymbol{\phi}_{ij}),
\qquad
\boldsymbol{\xi}_{ij}\triangleq
\mathbf{J}^R_{ij}\delta \mathbf{b}_g,
\label{eq:cdeltas}
\end{equation}
where $\delta\boldsymbol{\phi}_{ij}\in\mathbb R^3$ is the accumulated
rotation error.

Setting $\delta\boldsymbol{\phi}_{ij}=\mathbf 0$ in~\eqref{eq:cdeltas} gives
the noise-free rotation relation
\begin{equation}
\mathbf{R}_j
=
\mathbf{R}_i\Delta\bar{\mathbf{R}}_{ij}\,
\Cay(\boldsymbol{\xi}_{ij})
=
\mathbf{R}_i\Delta\bar{\mathbf{R}}_{ij}
\bigl(\mathbf{I}_3-\tfrac12\boldsymbol{\xi}_{ij}\wed\bigr)^{-1}
\bigl(\mathbf{I}_3+\tfrac12\boldsymbol{\xi}_{ij}\wed\bigr).
\label{eq:caypred}
\end{equation}
For any $\boldsymbol{\xi}_{ij}\in\R^3$,
\begin{equation}
\det\bigl(\mathbf{I}_3-\tfrac12\boldsymbol{\xi}_{ij}\wed\bigr)
=
1+\tfrac14\|\boldsymbol{\xi}_{ij}\|^2
>
0,
\label{eq:caydet}
\end{equation}
so the denominator is always invertible. Moreover,
$\mathbf{I}_3-\tfrac12\boldsymbol{\xi}_{ij}\wed$ and
$\mathbf{I}_3+\tfrac12\boldsymbol{\xi}_{ij}\wed$ commute because both are
polynomials in $\boldsymbol{\xi}_{ij}\wed$. We can therefore right-multiply
\eqref{eq:caypred} by the denominator to obtain
\begin{equation}
\mathbf{R}_j
\bigl(\mathbf{I}_3-\tfrac12\boldsymbol{\xi}_{ij}\wed\bigr)
=
\mathbf{R}_i\Delta\bar{\mathbf{R}}_{ij}
\bigl(\mathbf{I}_3+\tfrac12\boldsymbol{\xi}_{ij}\wed\bigr).
\label{eq:caycleared}
\end{equation}
This gives the polynomial rotation residual
\begin{equation}
\mathbf{r}^R_{ij}
\triangleq
\bigl(\mathbf{R}_j-\mathbf{R}_i\Delta\bar{\mathbf{R}}_{ij}\bigr)
-
\tfrac12
\bigl(\mathbf{R}_j+\mathbf{R}_i\Delta\bar{\mathbf{R}}_{ij}\bigr)
\boldsymbol{\xi}_{ij}\wed.
\label{eq:rRpoly}
\end{equation}
Unlike the geodesic residual $\mathbf r^\phi_{ij}$ in~\eqref{eq:resphi},
\eqref{eq:rRpoly} contains neither the logarithmic map nor trigonometric
functions. Matrix rotation residuals of this form are commonly used in
certifiable state estimation~\cite{sesync,cora}. The remaining
rotation--bias products are handled through lifting.

\subsection{Lifting to the QCQP Formulation}
\label{sec:map}

For an IMU pre-integration edge $(i,j)$, the Cayley-based factor is
\begin{equation}
F^{\Cay}_{ij}
=
\tau_{R,ij}\left\|\mathbf{r}^R_{ij}\right\|_F^2
+
\tau_{v,ij}\left\|\mathbf{r}^v_{ij}\right\|_2^2
+
\tau_{p,ij}\left\|\mathbf{r}^p_{ij}\right\|_2^2,
\label{eq:cayfactor}
\end{equation}
where the scalar precisions $\tau_{R,ij}$, $\tau_{v,ij}$, and
$\tau_{p,ij}$ are defined below. The residuals are
\begin{subequations}
\begin{equation}
\mathbf{r}^R_{ij}
=
\left(\mathbf{R}_j-\mathbf{R}_i\Delta\bar{\mathbf{R}}_{ij}\right)
-
\frac{1}{2}
\left(\mathbf{R}_j+\mathbf{R}_i\Delta\bar{\mathbf{R}}_{ij}\right)
\boldsymbol{\xi}_{ij}\wed,
\label{eq:cayfactor_rot}
\end{equation}

\begin{align}
\mathbf{r}^v_{ij}
&=
\mathbf{R}_i^\top
\left(
\mathbf{v}_j-\mathbf{v}_i-\mathbf{g}^{\mathrm W}\Delta t_{ij}
\right) \notag\\
&\qquad
-\Delta\bar{\mathbf{v}}_{ij}
-\mathbf{A}^{vg}_{ij}\delta \mathbf{b}_g
-\mathbf{A}^{va}_{ij}\delta \mathbf{b}_a,
\label{eq:cayfactor_v}\\
\mathbf{r}^p_{ij}
&=
\mathbf{R}_i^\top
\left(
\mathbf{p}_j-\mathbf{p}_i-\mathbf{v}_i\Delta t_{ij}
-\tfrac12 \mathbf{g}^{\mathrm W}\Delta t_{ij}^2
\right) \notag\\
&\qquad
-\Delta\bar{\mathbf{p}}_{ij}
-\mathbf{A}^{pg}_{ij}\delta \mathbf{b}_g
-\mathbf{A}^{pa}_{ij}\delta \mathbf{b}_a.
\label{eq:cayfactor_p}
\end{align}
\end{subequations}
Here, $\mathbf A^{vg}_{ij}$, $\mathbf A^{va}_{ij}$,
$\mathbf A^{pg}_{ij}$, and $\mathbf A^{pa}_{ij}\in\mathbb R^{3\times3}$
are the accumulated Jacobians of the Cayley pre-integrated velocity and
position with respect to the gyroscope and accelerometer biases. All
Jacobians are evaluated at the nominal bias.

Although the residuals are now polynomial, some terms still contain products
of unknowns. The rotation residual contains rotation--bias products, while the
translation residuals contain rotation--state and rotation--bias products.
We introduce auxiliary variables for these terms so that each residual becomes
affine in the lifted variables.

\paragraph{Rotation--Bias Products}
Using
$\delta\mathbf{b}_g=\mathbf{b}_g-\bar{\mathbf{b}}_g$, define
\begin{subequations}
\begin{equation}
\boldsymbol{\Gamma}_{ij}
\triangleq
\left(\mathbf{J}^R_{ij}\bar{\mathbf{b}}_g\right)\wed,
\qquad
\breve{\mathbf{R}}_{ij}
\triangleq
\Delta\bar{\mathbf{R}}_{ij}
\left(\mathbf{I}_3-\frac12\boldsymbol{\Gamma}_{ij}\right).
\end{equation}
The rotation residual can then be written as
\begin{equation}
\mathbf{r}^R_{ij}
=
\mathbf{R}_j
\left(\mathbf{I}_3+\frac12\boldsymbol{\Gamma}_{ij}\right)
-
\mathbf{R}_i\breve{\mathbf{R}}_{ij}
-
\frac12\left(\mathbf{Y}_{ij}+\mathbf Y'_{ij}\right),
\label{eq:rR}
\end{equation}
where
\begin{equation}
\mathbf{Y}_{ij}
\triangleq
\mathbf{R}_i\Delta\bar{\mathbf{R}}_{ij}
\left(\mathbf{J}^R_{ij}\mathbf{b}_g\right)\wed,
\qquad
\mathbf Y'_{ij}
\triangleq
\mathbf{R}_j
\left(\mathbf{J}^R_{ij}\mathbf{b}_g\right)\wed.
\label{eq:Ylift}
\end{equation}
\end{subequations}
The matrices $\boldsymbol{\Gamma}_{ij}$ and
$\breve{\mathbf R}_{ij}$ are known. We introduce
$\mathbf Y_{ij}$ and $\mathbf Y'_{ij}$ as auxiliary variables, making
$\mathbf r^R_{ij}$ affine. Their definitions are enforced by quadratic
constraints below.

\paragraph{Rotation--State Products and Homogenization}
The translation residuals in~\eqref{eq:cayfactor_v} and
\eqref{eq:cayfactor_p} contain products such as
$\mathbf R_i^\top\mathbf v_j$ and
$\mathbf R_i^\top\mathbf p_j$. Rather than lifting these products directly,
we use
\[
\tau\|\mathbf r\|_2^2
=
\tau\|\mathbf R_i\mathbf r\|_2^2,
\qquad
\mathbf R_i\in\SO(3).
\]
This equality requires an isotropic weight. We therefore approximate the three
covariance blocks by scalar precisions
\begin{equation}
\tau_{R,ij}^{-1}
=
\tfrac13\tr\boldsymbol{\Sigma}^\phi_{ij},
\quad
\tau_{v,ij}^{-1}
=
\tfrac13\tr\boldsymbol{\Sigma}^v_{ij},
\quad
\tau_{p,ij}^{-1}
=
\tfrac13\tr\boldsymbol{\Sigma}^p_{ij},
\label{eq:iso}
\end{equation}
where $\boldsymbol{\Sigma}^\phi_{ij}$,
$\boldsymbol{\Sigma}^v_{ij}$, and
$\boldsymbol{\Sigma}^p_{ij}$ are the diagonal covariance blocks of
$\boldsymbol{\Sigma}^{\mathrm f}_{ij}$. This approximation drops the
cross-correlations and directional weighting in the original covariance. Its
effect is evaluated in Sec.~\ref{sec:cmp}.

Premultiplying the translation residuals by $\mathbf R_i$ removes the
rotation--state products. We also introduce a homogenizing scalar $s$ satisfying
$s^2=1$. Define
\begin{subequations}
\begin{align}
\breve{\mathbf{v}}_{ij}
&\triangleq
\Delta\bar{\mathbf{v}}_{ij}
-
\mathbf{A}^{vg}_{ij}\bar{\mathbf{b}}_g
-
\mathbf{A}^{va}_{ij}\bar{\mathbf{b}}_a,\\
\breve{\mathbf{p}}_{ij}
&\triangleq
\Delta\bar{\mathbf{p}}_{ij}
-
\mathbf{A}^{pg}_{ij}\bar{\mathbf{b}}_g
-
\mathbf{A}^{pa}_{ij}\bar{\mathbf{b}}_a.
\end{align}
\end{subequations}
Using $\operatorname{vec}(\cdot)$ for column-wise vectorization and
$\otimes$ for the Kronecker product, the transformed residuals are
\begin{subequations}
\begin{align}
\mathbf{R}_i \mathbf{r}^v_{ij}
&=
\left(
\mathbf{v}_j-\mathbf{v}_i-\mathbf{g}^{\mathrm W}\Delta t_{ij}s
\right) \notag\\
&\qquad
-\left(\breve{\mathbf{v}}_{ij}^{\top}\otimes \mathbf{I}_3\right)
\operatorname{vec}(\mathbf{R}_i)
-\mathbf{W}^v_{ij},
\label{eq:rv}\\
\mathbf{R}_i \mathbf{r}^p_{ij}
&=
\left(
\mathbf{p}_j-\mathbf{p}_i-\mathbf{v}_i\Delta t_{ij}
-\tfrac12 \mathbf{g}^{\mathrm W}\Delta t_{ij}^2s
\right) \notag\\
&\qquad
-\left(\breve{\mathbf{p}}_{ij}^{\top}\otimes \mathbf{I}_3\right)
\operatorname{vec}(\mathbf{R}_i)
-\mathbf{W}^p_{ij},
\label{eq:rpq}
\end{align}
\end{subequations}
where
\begin{subequations}
\begin{align}
\mathbf{W}^v_{ij}
&\triangleq
\mathbf{R}_i \mathbf{u}^v_{ij},
&
\mathbf{u}^v_{ij}
&\triangleq
\mathbf{A}^{vg}_{ij}\mathbf{b}_g
+\mathbf{A}^{va}_{ij}\mathbf{b}_a,
\label{eq:Wlift}\\
\mathbf{W}^p_{ij}
&\triangleq
\mathbf{R}_i \mathbf{u}^p_{ij},
&
\mathbf{u}^p_{ij}
&\triangleq
\mathbf{A}^{pg}_{ij}\mathbf{b}_g
+\mathbf{A}^{pa}_{ij}\mathbf{b}_a.
\label{eq:Wplift}
\end{align}
\end{subequations}
The vectors $\mathbf W^v_{ij}$ and $\mathbf W^p_{ij}$ represent the remaining
rotation--bias products. Their definitions are also enforced by quadratic
constraints.

\paragraph{Vectorization}
We use the Kronecker identity~\cite{horn}
\begin{subequations}
\begin{equation}
\operatorname{vec}(\mathbf A\mathbf B\mathbf C)
=
(\mathbf C^\top\otimes\mathbf A)
\operatorname{vec}(\mathbf B).
\label{eq:vecidentity}
\end{equation}
In particular, for any $\mathbf u\in\mathbb R^3$,
\begin{equation}
\mathbf{R}_i\mathbf u
=
\left(\mathbf{u}^\top\otimes \mathbf{I}_3\right)
\operatorname{vec}(\mathbf{R}_i).
\label{eq:vecRu}
\end{equation}
\end{subequations}
The three residuals are therefore affine in the original and auxiliary
variables.

\paragraph{Assembly}
For an edge $(i,j)$, $\mathbf b_g$ and $\mathbf b_a$ denote the bias variables
at keyframe $i$, with their keyframe indices omitted for readability. Let
$\mathbf b_m=[\mathbf b_{g,m}^\top\ \mathbf b_{a,m}^\top]^\top$ denote the
stacked bias at keyframe $m$. We collect the state variables, biases,
auxiliary variables, and homogenizing scalar into
\begin{equation}
\begin{aligned}
\mathbf{x} \triangleq \Big[\
&\{\operatorname{vec}(\mathbf{R}_i),\mathbf{v}_i,\mathbf{p}_i\}_i,\;
\{\mathbf{b}_m\}_m,\\
&\{\operatorname{vec}(\mathbf{Y}_{ij}),
\operatorname{vec}(\mathbf{Y}'_{ij}),
\mathbf{W}^v_{ij},\mathbf{W}^p_{ij}\}_{(i,j)},\; s\ \Big]^\top .
\end{aligned}
\label{eq:decvec}
\end{equation}
Here, $i$ and $m$ index the keyframes, while $(i,j)$ indexes the
pre-integration edges.

Since every residual is affine in $\mathbf x$, each edge contributes a
quadratic cost
\begin{equation}
\|\mathbf{L}_{ij}\mathbf{x}\|_2^2
=
\mathbf{x}^\top \mathbf{L}_{ij}^\top \mathbf{L}_{ij}\mathbf{x}
=
\left\langle \mathbf{Q}_{ij},\mathbf{x}\mathbf{x}^\top\right\rangle,
\qquad
\mathbf{Q}_{ij}\succeq0.
\label{eq:qform}
\end{equation}
Here, $\mathbf L_{ij}$ collects the weighted residual coefficients and
$\mathbf Q_{ij}\triangleq\mathbf L_{ij}^\top\mathbf L_{ij}$.

The auxiliary variables are linked to their corresponding products through
the quadratic constraints
\begin{subequations}
\label{eq:liftdef}
\begin{align}
\left(\mathbf{R}_i\Delta\bar{\mathbf{R}}_{ij}\right)^\top
\mathbf{Y}_{ij}
&=
\left(\mathbf{J}^R_{ij}\mathbf{b}_g\right)\wed s,
\label{eq:Yconstraint}\\
\mathbf{R}_j^\top\mathbf Y'_{ij}
&=
\left(\mathbf{J}^R_{ij}\mathbf{b}_g\right)\wed s,
\label{eq:Yprimeconstraint}\\
\mathbf{R}_i^\top \mathbf{W}^v_{ij}
&=
\mathbf{u}^v_{ij}s,
\qquad
\mathbf{R}_i^\top \mathbf{W}^p_{ij}
=
\mathbf{u}^p_{ij}s.
\label{eq:Wconstraints}
\end{align}
\end{subequations}
The remaining primary constraints are the homogenizing scalar constraint and the rotation constraints~\cite{TRC15}:
\begin{equation}
s^2=1,
\quad
\mathbf{R}_i^\top \mathbf{R}_i=s^2\mathbf{I}_3,
\quad
\mathbf{R}_i^{(a)}\times\mathbf{R}_i^{(b)}=s\mathbf{R}_i^{(c)},
\label{eq:manifold}
\end{equation}
where $\mathbf{R}^{(k)}$ denotes the $k$-th column of $\mathbf{R}$ and $(a,b,c)$ runs over the cyclic permutations of $(1,2,3)$.
Together,~\eqref{eq:qform}, \eqref{eq:liftdef}, and
\eqref{eq:manifold} provide the quadratic cost and primary quadratic
constraints needed to include the proposed factor in a QCQP.

\section{Tightening the SDP Relaxation}
\label{sec:redundant}

Using a QCQP-representable IMU factor in certifiable estimation requires
the corresponding convex relaxation to be sufficiently tight, either
naturally~\cite{sesync} or through additional redundant
constraints~\cite{autotight}. For the proposed factor, applying the standard
SDP relaxation to the primary constraints of Sec.~\ref{sec:map} generally
produces high-rank solutions. We therefore derive additional quadratic
identities to tighten the relaxation.
\footnote{
Redundant constraints are identities satisfied by every feasible solution
of the original problem. They do not change its global optimum, but can
strengthen its convex relaxation. See~\cite{autotight} for a detailed
discussion of their construction}

Following the AutoTight approach~\cite{autotight}, we organize these
identities into three groups: rotation constraints, matrix-valued wedge
lifting constraints, and vector-valued translation lifting constraints.
Together, these groups contain the four subsets A1, B1, B2, and C1 used in the experiments.
\paragraph{Rotation Constraints}
Note that the primary constraints in~\eqref{eq:manifold} include column orthonormality, $\mathbf{R}_k^\top\mathbf{R}_k=s^2\mathbf{I}_3$. For every feasible rotation, column orthonormality also implies row orthonormality. We therefore add
\begin{equation}\label{eq:catalog}
    \mathbf{R}_k\mathbf{R}_k^\top=s^2\mathbf{I}_3 \quad (\mathrm{A1},\,6/\text{node}).
\end{equation}

\paragraph{Wedge Lifting Constraints}
Let $\mathcal{E}_{\mathrm{IMU}}$ denote the set of IMU pre-integration
edges. For each $(i,j)\in\mathcal{E}_{\mathrm{IMU}}$, define
\[
\mathbf{M}_{ij}
\triangleq
\mathbf{R}_i\Delta\bar{\mathbf{R}}_{ij},
\qquad
\mathbf{w}_{ij}
\triangleq
\mathbf{J}^R_{ij}\bvec_{g,m(i,j)},
\]
where $m(i,j)$ denotes the bias block assigned to edge $(i,j)$. From the
definition of the wedge lifting,
\[
\mathbf{Y}_{ij}
=
\mathbf{M}_{ij}\mathbf{w}_{ij}\wed.
\]
Since
$\mathbf{M}_{ij}^\top\mathbf{M}_{ij}=\mathbf{I}_3$,
standard skew-symmetric identities give
\begin{align}
\mathbf{Y}_{ij}\mathbf{w}_{ij}
&=
\mathbf{M}_{ij}
(\mathbf{w}_{ij}\times\mathbf{w}_{ij})
=
\mathbf{0}
\quad
(\mathrm{B1},\,3/\text{lifting}),
\label{eq:B1}\\
\mathbf{Y}_{ij}^\top\mathbf{Y}_{ij}
&=
(\mathbf{w}_{ij}\wed)^\top\mathbf{w}_{ij}\wed
=
\|\mathbf{w}_{ij}\|^2\mathbf{I}_3
-
\mathbf{w}_{ij}\mathbf{w}_{ij}^\top
\quad
(\mathrm{B2}).
\label{eq:B2}
\end{align}
B1 gives a nullspace relation between the lifted matrix and the
bias-dependent vector, while B2 fixes the Gram matrix of the lifting.
The same two identities are applied to
$\mathbf{Y}'_{ij}$, defined in~\eqref{eq:Ylift}. Thus, B1 contains six
scalar constraints per edge, three for each matrix lifting.

\paragraph{Vector Lifting Constraints}
The translation liftings in
\eqref{eq:Wlift}--\eqref{eq:Wplift} satisfy
\[
\mathbf{W}^v_{ij}
=
\mathbf{R}_i\mathbf{u}^v_{ij},
\qquad
\mathbf{W}^p_{ij}
=
\mathbf{R}_i\mathbf{u}^p_{ij}.
\]
Using the orthogonality of $\mathbf{R}_i$, the velocity lifting satisfies
\begin{equation}
\|\mathbf{W}^v_{ij}\|^2
=
(\mathbf{u}^v_{ij})^\top
\mathbf{R}_i^\top\mathbf{R}_i
\mathbf{u}^v_{ij}
=
\|\mathbf{u}^v_{ij}\|^2
\quad
(\mathrm{C1},\,2/\text{edge}).
\label{eq:C1}
\end{equation}
The same scalar identity is applied to the position lifting,
\[
\|\mathbf{W}^p_{ij}\|^2
=
\|\mathbf{u}^p_{ij}\|^2.
\]
Finally, Table~\ref{tab:catalog} summarizes the four constraint subsets.

\begin{table}[t]
\centering
\caption{Redundant-constraint subsets used to tighten the SDP relaxation.
Here, $\mathbf{Y}_{ij}^{(\prime)}$ denotes either
$\mathbf{Y}_{ij}$ or $\mathbf{Y}'_{ij}$, and
$\bullet\in\{v,p\}$.}
\label{tab:catalog}
\small
\setlength{\tabcolsep}{4pt}
\begin{tabular}{@{}lll@{}}
\toprule
\textbf{ID} & \textbf{Identity} & \textbf{Constrains} \\
\midrule
A1
& $\mathbf{R}_k\mathbf{R}_k^\top=s^2\mathbf{I}_3$
& rotation rows \\
B1
& $\mathbf{Y}_{ij}^{(\prime)}\mathbf{w}_{ij}=\mathbf{0}$
& lifting--bias \\
B2
& $(\mathbf{Y}_{ij}^{(\prime)})^\top
\mathbf{Y}_{ij}^{(\prime)}
=
\|\mathbf{w}_{ij}\|^2\mathbf{I}_3
-
\mathbf{w}_{ij}\mathbf{w}_{ij}^\top$
& lifting--lifting \\
C1
& $\|\mathbf{W}^\bullet_{ij}\|^2
=
\|\mathbf{u}^\bullet_{ij}\|^2$
& vector lifting \\
\bottomrule
\end{tabular}
\end{table}

\section{GNSS-Aided Inertial Smoothing}
\label{sec:problem}

This section considers GNSS-aided inertial smoothing as an application of the
proposed QCQP-representable IMU pre-integration factor. The problem combines
IMU measurements with GNSS position and velocity measurements using a
relatively simple sensor setup. Conventional estimators typically rely on
local nonlinear optimization alone. With the factor developed in
Sec.~\ref{sec:preint}, we can apply certifiable methods to the same problem.
We first define the problem and its factor-graph representation, and then show
how the QCQP formulation and constraints developed in the previous sections
apply.

\subsection{Problem Setup}
\label{sec:problemsetup}

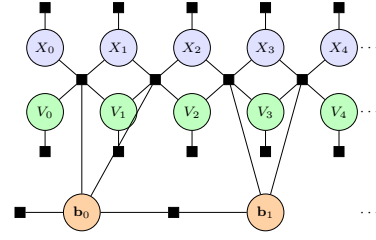
\begin{figure}[t]
\centering
\begin{tikzpicture}[
  scale=0.736, transform shape,
  var/.style={circle,draw,minimum size=6.6mm,inner sep=0pt,font=\scriptsize},
  fac/.style={rectangle,draw,fill=black,minimum size=1.8mm,inner sep=0pt},
  every path/.style={line width=0.35pt}]

\foreach \i in {0,1,2,3,4}{
  \node[var,fill=blue!12]    (X\i) at (1.32*\i, 2.10) {$X_{\i}$};
  \node[var,fill=green!25]  (V\i) at (1.32*\i, 0.95) {$V_{\i}$};
  \node[fac] (fx\i) at (1.32*\i, 2.82) {}; \draw (fx\i) -- (X\i);
  \node[fac] (fv\i) at (1.32*\i, 0.24) {}; \draw (fv\i) -- (V\i);
}

\node[var,fill=orange!35] (B0) at (0.66,-0.85) {$\mathbf{b}_{0}$};
\node[var,fill=orange!35] (B1) at (3.96,-0.85) {$\mathbf{b}_{1}$};
\node[fac] (fb) at (-0.45,-0.85) {}; \draw (fb) -- (B0);
\node[fac] (frw) at (2.31,-0.85) {};
\draw (B0) -- (frw); \draw (frw) -- (B1);

\foreach \i/\j in {0/1,1/2}{
  \node[fac] (fi\i) at (1.32*\i+0.66, 1.53) {};
  \draw (fi\i) -- (X\i); \draw (fi\i) -- (X\j);
  \draw (fi\i) -- (V\i); \draw (fi\i) -- (V\j);
  \draw (fi\i) -- (B0);
}
\foreach \i/\j in {2/3,3/4}{
  \node[fac] (fi\i) at (1.32*\i+0.66, 1.53) {};
  \draw (fi\i) -- (X\i); \draw (fi\i) -- (X\j);
  \draw (fi\i) -- (V\i); \draw (fi\i) -- (V\j);
  \draw (fi\i) -- (B1);
}
\node[font=\scriptsize] at (5.85, 2.10) {$\cdots$};
\node[font=\scriptsize] at (5.85, 0.95) {$\cdots$};
\node[font=\scriptsize] at (5.85,-0.85) {$\cdots$};
\end{tikzpicture}
\caption{GNSS--IMU smoothing factor graph, drawn for $M=2$ bias blocks of two edges each.
Circles are variables, squares are factors. Each keyframe state is split into a pose part
$X_i=(\mathbf{R}_i,\mathbf{p}_i)$ (blue) and a velocity part $V_i=\mathbf{v}_i$ (green); the
top and bottom factors are the GNSS position and Doppler-velocity measurements~\eqref{eq:gnss},
present at every keyframe. Each inertial factor $F^{\Cay}_{ij}$~\eqref{eq:cayfactor} couples
the adjacent poses and velocities $X_i,V_i,X_j,V_j$ to the bias block $\mathbf{b}_m$ of its edge
(orange). $\mathbf{b}_0$ carries the prior~\eqref{eq:biasprior}; consecutive blocks are joined
by the random-walk factor~\eqref{eq:biasrw}, which informs
$\mathbf{b}_1,\dots,\mathbf{b}_{M-1}$.}
\label{fig:fg}
\end{figure}

Over $K$ keyframes and $M$ bias blocks, we estimate
\begin{equation}
\mathcal X
\triangleq
\{\mathbf{R}_i,\mathbf{v}_i,\mathbf{p}_i\}_{i=0}^{K-1}
\cup
\{\mathbf{b}_m\}_{m=0}^{M-1},
\label{eq:states}
\end{equation}
where $\mathbf R_i\in\SO(3)$ and
$\mathbf v_i,\mathbf p_i\in\mathbb R^3$ are the orientation, velocity, and
position at keyframe $i$. Each bias block is
$\mathbf b_m
=[\mathbf b_{g,m}^{\top}\ \mathbf b_{a,m}^{\top}]^{\top}
\in\mathbb R^6$.
Let $E$ denote the set of IMU pre-integration edges. We partition $E$ into
the disjoint sets $E_0,\ldots,E_{M-1}$, where all edges in $E_m$ share the
bias $\mathbf b_m$. The lifted decision vector $\mathbf x$
in~\eqref{eq:decvec} augments $\mathcal X$ with the auxiliary variables and
homogenizing scalar introduced in Sec.~\ref{sec:map}.
At the nominal bias $\bar{\mathbf b}$, the factors in Fig.~\ref{fig:fg}
follow the measurement models
\begin{subequations}
\label{eq:genmodel}
\begin{align}
\Delta\bar{\mathbf{R}}_{ij}
&=
\mathbf{R}_i^{\top}\mathbf{R}_j\,
\Cay(\delta\boldsymbol{\phi}_{ij}),
\label{eq:gm-imu}\\
\Delta\bar{\mathbf{v}}_{ij}
&=
\mathbf{R}_i^{\top}
\bigl(
\mathbf{v}_j-\mathbf{v}_i-\mathbf{g}^{\W}\Delta t_{ij}
\bigr)
+\delta\mathbf{v}_{ij},
\notag\\
\Delta\bar{\mathbf{p}}_{ij}
&=
\mathbf{R}_i^{\top}
\bigl(
\mathbf{p}_j-\mathbf{p}_i-\mathbf{v}_i\Delta t_{ij}
-\tfrac12\mathbf{g}^{\W}\Delta t_{ij}^2
\bigr)
+\delta\mathbf{p}_{ij},
\notag\\
\mathbf{z}^{p}_i
&=
\mathbf{p}_i+\boldsymbol{\varepsilon}^{p}_i,
\qquad
\mathbf{z}^{v}_i
=
\mathbf{v}_i+\boldsymbol{\varepsilon}^{v}_i,
\label{eq:gnss}\\
\mathbf{b}_0
&=
\bar{\mathbf{b}}+\boldsymbol{\varepsilon}^{b},
\label{eq:biasprior}\\
\mathbf{b}_{m+1}
&=
\mathbf{b}_m+\boldsymbol{\varepsilon}^{\mathrm{rw}}_m.
\label{eq:biasrw}
\end{align}
\end{subequations}
Under the isotropic approximation of~\eqref{eq:iso}, we assume that the
factor noises are independent and zero-mean Gaussian:
\begin{equation}
\begin{aligned}
\delta\boldsymbol{\phi}_{ij}
&\sim
\mathcal N\bigl(
\mathbf{0},\tau_{R,ij}^{-1}\mathbf{I}_3
\bigr),
&
\delta\mathbf{v}_{ij}
&\sim
\mathcal N\bigl(
\mathbf{0},\tau_{v,ij}^{-1}\mathbf{I}_3
\bigr),\\
\delta\mathbf{p}_{ij}
&\sim
\mathcal N\bigl(
\mathbf{0},\tau_{p,ij}^{-1}\mathbf{I}_3
\bigr),
&
\boldsymbol{\varepsilon}^{p}_i
&\sim
\mathcal N\bigl(
\mathbf{0},\tau_{\pi p}^{-1}\mathbf{I}_3
\bigr),\\
\boldsymbol{\varepsilon}^{v}_i
&\sim
\mathcal N\bigl(
\mathbf{0},\tau_{\pi v}^{-1}\mathbf{I}_3
\bigr),
&
\boldsymbol{\varepsilon}^{b}
&\sim
\mathcal N\bigl(
\mathbf{0},\boldsymbol{\Lambda}^{-1}
\bigr),\\
\boldsymbol{\varepsilon}^{\mathrm{rw}}_m
&\sim
\mathcal N\bigl(
\mathbf{0},\boldsymbol{\Lambda}_{\mathrm{rw}}^{-1}
\bigr).
\end{aligned}
\label{eq:noises}
\end{equation}

The first three relations in~\eqref{eq:genmodel} use the Cayley
pre-integrated measurements from Sec.~\ref{sec:cayint}. The triple
$(\Delta\bar{\mathbf R}_{ij},\Delta\bar{\mathbf v}_{ij},
\Delta\bar{\mathbf p}_{ij})$ is computed once at the nominal bias
$\bar{\mathbf b}$. Changes in the estimated bias are handled by the rotation
correction in~\eqref{eq:cdeltas} and the bias Jacobians in
\eqref{eq:cayfactor_v}--\eqref{eq:cayfactor_p}. The precisions
$\tau_{R,ij}$, $\tau_{v,ij}$, and $\tau_{p,ij}$ are defined
in~\eqref{eq:iso}.

The measurements $\mathbf z_i^p$ and $\mathbf z_i^v$ are the GNSS position
and velocity measurements, with scalar precisions $\tau_{\pi p}$ and
$\tau_{\pi v}$. We use the nominal bias $\bar{\mathbf b}$ as the mean of the
initial bias prior in~\eqref{eq:biasprior}, with
\[
\boldsymbol{\Lambda}
=
\mathrm{diag}
\bigl(
\tau_{\pi bg}\mathbf I_3,
\tau_{\pi ba}\mathbf I_3
\bigr),
\]
where $\tau_{\pi bg}$ and $\tau_{\pi ba}$ are the gyroscope- and
accelerometer-bias prior precisions. The random-walk model
in~\eqref{eq:biasrw} uses
\[
\boldsymbol{\Lambda}_{\mathrm{rw}}
=
\mathrm{diag}
\bigl(
\sigma_{bg}^{-2}\Delta t_b^{-1}\mathbf I_3,
\sigma_{ba}^{-2}\Delta t_b^{-1}\mathbf I_3
\bigr),
\]
where $\sigma_{bg}$ and $\sigma_{ba}$ are the bias random-walk noise
parameters and $\Delta t_b$ is the time between consecutive bias blocks.

\subsection{Applying Certifiable Methods}
\label{sec:certest}

Under the Gaussian noise models in~\eqref{eq:noises}, the negative
log-posterior is a sum of squared residuals, giving the following estimator.

\begin{problem}[GNSS-aided inertial smoothing]
\label{prob:map}
Given pre-integrated inertial measurements
$\{\Delta\bar{\mathbf{R}}_{ij},
\Delta\bar{\mathbf{v}}_{ij},
\Delta\bar{\mathbf{p}}_{ij}\}_{(i,j)\in E}$
and GNSS observations
$\{\mathbf{z}^{p}_i,\mathbf{z}^{v}_i\}_{i=0}^{K-1}$, estimate
$\mathcal X$ by solving
\begin{equation}
\begin{aligned}
\min_{\mathcal X}\ \ 
&\sum_{(i,j)\in E}F^{\Cay}_{ij}
+\sum_i
\Bigl(
\tau_{\pi p}\|\mathbf{p}_i-\mathbf{z}^{p}_i\|^2
+\tau_{\pi v}\|\mathbf{v}_i-\mathbf{z}^{v}_i\|^2
\Bigr)\\
&+\sum_{m=0}^{M-2}
\|\mathbf{b}_{m+1}-\mathbf{b}_m\|^2_
{\boldsymbol{\Lambda}_{\mathrm{rw}}}
+\|\mathbf{b}_0-\bar{\mathbf{b}}\|^2_{\boldsymbol{\Lambda}}
\end{aligned}
\label{eq:map}
\end{equation}
where $F^{\Cay}_{ij}$ is the Cayley inertial factor
in~\eqref{eq:cayfactor}. For $(i,j)\in E_m$, this factor uses the bias
$\mathbf b_m$.
\end{problem}

After applying the liftings of Sec.~\ref{sec:map},
Problem~\ref{prob:map} admits a QCQP representation. The inertial factors are
quadratic in the lifted variables, while the GNSS and bias terms are squared
affine functions of the state variables. Using the homogenizing scalar $s$,
these terms can be written as
\begin{equation}
\|\mathbf{p}_i-\mathbf{z}^{p}_i s\|^2,\quad
\|\mathbf{b}_{m+1}-\mathbf{b}_m\|^2_
{\boldsymbol{\Lambda}_{\mathrm{rw}}},\quad
\|\mathbf{b}_0-\bar{\mathbf{b}}s\|^2_{\boldsymbol{\Lambda}}.
\notag
\end{equation}
The GNSS velocity term is handled in the same way. Collecting all factor costs
into a single matrix $\mathbf Q\succeq0$ gives
\[
\min_{\mathbf{x}}
\left\langle
\mathbf Q,\mathbf{x}\mathbf{x}^{\top}
\right\rangle
\]
subject to the primary constraints in~\eqref{eq:liftdef} and
\eqref{eq:manifold}, together with the redundant constraints from
Sec.~\ref{sec:redundant}. We apply certifiable methods to this QCQP
formulation.

\section{Experiments}
\label{sec:exp}

We evaluate the proposed pre-integration factor and redundant constraints on
synthetic and real-world GNSS--IMU smoothing problems. The experiments address
three questions: (i) which subsets of redundant constraints are needed to
obtain a tight SDP relaxation, and how tightness changes with measurement
noise; (ii) whether the relaxation remains tight on real-world data across
different window sizes and bias partitions; and (iii) how the isotropic
weighting used in Sec.~\ref{sec:preint} affects estimation accuracy compared
with standard anisotropic pre-integration. All SDPs are solved using MOSEK on
an Intel i7-8700K CPU. Keyframes are sampled at $1\,\mathrm{Hz}$, and the bias
is divided into $M\in\{1,2,3\}$ blocks.

\subsection{Experimental Setup}

\paragraph{Evaluation Metrics}
Let $\hat{\mathbf Z}$ denote the optimal SDP solution. We compute its numerical
rank as the number of eigenvalues larger than
$10^{-6}\lambda_{\max}(\hat{\mathbf Z})$ and regard the relaxation as
numerically tight when $\rank(\hat{\mathbf Z})=1$. We also report the relative
optimality gap
$\eta
\triangleq
\frac{
\left|
\hat{\mathbf x}^{\top}\mathbf Q\hat{\mathbf x}-p^\star
\right|
}{
\left|
\hat{\mathbf x}^{\top}\mathbf Q\hat{\mathbf x}
\right|
},
$
where $p^\star$ is the SDP dual optimum and $\hat{\mathbf x}$ is the recovered
point. To recover $\hat{\mathbf x}$, we project the rotation blocks onto
$\SO(3)$ using SVD and rebuild the lifting variables from
\eqref{eq:Ylift}--\eqref{eq:Wplift}. A rank-one solution with a negligible
optimality gap verifies global optimality to numerical precision. We measure
estimation accuracy using the mean geodesic orientation error and mean
Euclidean position error with respect to ground truth.

\paragraph{Benchmarks}
We generate synthetic trajectories from two constant-twist motion primitives:
\texttt{sick}~\cite{gtsam}, which provides roll--pitch excitation, and
\texttt{loop}~\cite{gtsam}, which combines forward motion with yaw. IMU
measurements are generated at $100\,\mathrm{Hz}$. The main ablation uses
$K=8$ keyframes, and we repeat the comparison with $K=16$.

The real-world experiments use the EuRoC MAV
\texttt{MH\_01\_easy} sequence~\cite{euroc}, which provides IMU measurements
at $200\,\mathrm{Hz}$, and the \texttt{medium urban} and \texttt{deep urban}
sequences from UrbanNav~\cite{urbannav}. Because EuRoC is an indoor dataset,
we simulate GNSS observations from its ground truth by adding measurement
noise. The selected UrbanNav windows contain weak excitation and long
stationary periods, making rotation estimation more difficult.

A \emph{window} is a contiguous trajectory segment used to form one smoothing
problem. With keyframes sampled at $1\,\mathrm{Hz}$, the time between the first
and last keyframes is $K-1$ seconds. Each window contains $K$ keyframes, its
bias is divided into $M$ blocks, and $n$ denotes the dimension of the decision
vector in~\eqref{eq:decvec}. We sample several windows for each configuration
$(K,M)$ using different starting offsets.

\begin{figure}[tb]
\centering
\includegraphics[width=0.8\columnwidth]{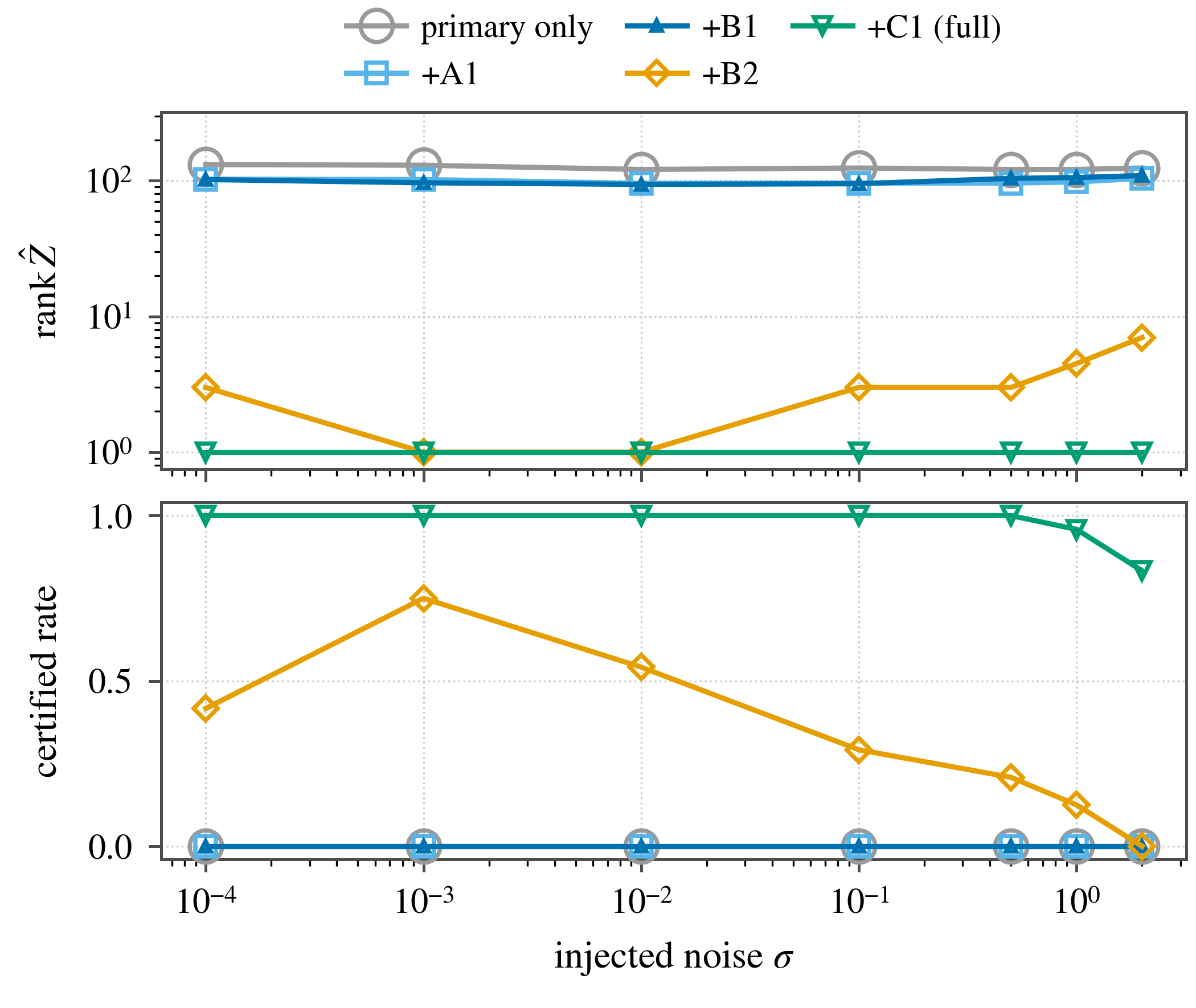}
\caption{\textbf{Effect of cumulative redundant-constraint subsets on SDP
tightness.} Numerical rank and certification rate over $840$ SDP solves.
Each stage includes all constraint subsets shown to its left. Adding B2
substantially tightens the relaxation, while the full set including C1
produces rank-one solutions across the tested noise range.}
\label{fig:ablation}
\end{figure}

\subsection{Ablation of Redundant Constraints}
\label{sec:ablation}

To address the first question, we construct synthetic problems with
$K=8$ keyframes and $M=2$ bias blocks, giving $n=340$. For each motion
primitive, we test seven noise levels and $12$ random seeds. 
The four
constraint sets are added cumulatively, giving
$2\times7\times12\times5=840$ SDP solves. For each solve, we record
$\rank(\hat{\mathbf Z})$ and the relative optimality gap $\eta$. The results
are shown in Fig.~\ref{fig:ablation}.

The primary constraints together with A1 and B1 are not sufficient to
produce a tight relaxation. Across the stages through B1, the relative
optimality gap remains at least $0.88$, and none of the corresponding $504$
solves is certified. Adding B2 substantially reduces both the numerical rank
and the optimality gap. With the full set, including C1, the median gap is
$2.8\times10^{-5}$, and the median numerical rank remains one across the
tested noise range, although the certification rate decreases at the highest
noise levels. We observe the same ordering for the larger $K=16$ problems
($n=652$).

\begin{table}[t]
\centering
\caption{\textbf{Performance on real-world benchmarks using the full
redundant-constraint set.} Four windows are evaluated for each configuration.
\textbf{Cert.} reports the number of rank-one certified windows. The median
and maximum relative gaps are reported in units of $10^{-6}$, and
\textbf{err.} is the median window-level mean orientation error.}
\label{tab:real}
\small
\setlength{\tabcolsep}{4pt}
\begin{tabular}{@{}llcccccc@{}}
\toprule
Sequence & $K$ & $M$ & $n$ & cert. & med~[$\eta$] & max~[$\eta$] & err.~[$^\circ$]\\
\midrule
EuRoC \texttt{MH\_01} & 8 & 1 & 334 & $4/4$ & 4.3 & 6.7 & 1.05 \\
& 16 & 1 & 646 & $4/4$ & 1.2 & 2.2 & 0.83 \\
& 16 & 2 & 652 & $4/4$ & 2.9 & 10 & 0.88 \\
& 24 & 1 & 958 & $4/4$ & 1.5 & 6.8 & 0.83 \\
& 24 & 3 & 970 & $4/4$ & 4.0 & 6.0 & 1.15 \\
\addlinespace
UrbanNav \texttt{med.} & 16 & 2 & 652 & $4/4$ & 9.0 & 39 & 0.72 \\
& 24 & 3 & 970 & $3/4$ & 340 & 640 & 0.30 \\
\addlinespace
UrbanNav \texttt{deep} & 16 & 2 & 652 & $4/4$ & 7.3 & 56 & 0.84 \\
& 24 & 3 & 970 & $4/4$ & 29 & 66 & 0.95 \\
\bottomrule
\end{tabular}
\end{table}

\subsection{Real-World Certifiability}
\label{sec:realworld}

To address the second question, we solve $36$ windows from EuRoC
\texttt{MH\_01} and the \texttt{medium urban} and \texttt{deep urban}
sequences of UrbanNav. The nine configurations range from $K=8$ to $24$
keyframes and use $M\in\{1,2,3\}$ bias blocks, with four windows per
configuration. For each window, we record $\rank(\hat{\mathbf Z})$, the
optimality gap $\eta$, and the mean orientation error. The results are
reported in Table~\ref{tab:real}.

The full constraint set produces rank-one certified solutions for $35$ of the
$36$ windows. The only exception is one $K=24$, $M=3$ window from
\texttt{medium urban}, for which the SDP solution has rank three. This window
contains a prolonged stationary segment, leaving the heading unobservable.
The resulting problem is degenerate and admits multiple equivalent solutions,
so the SDP may return a higher-rank combination of these solutions rather than
a single rank-one solution.

\begin{table}[t]
\centering
\caption{\textbf{Comparison with standard pre-integration}. Medians and
maxima are taken across the windows of each dataset; bold marks the best entry of a column.
Cert.\ counts windows returned with a rank-one certificate.}
\label{tab:cmp}
\small
\setlength{\tabcolsep}{4pt}
\begin{tabular}{@{}llccccc@{}}
\toprule
& & \multicolumn{2}{c}{rot.~[degree]} & \multicolumn{2}{c}{pos.~[m]} &\\
\cmidrule(lr){3-4}\cmidrule(lr){5-6}
Method & init & med & max & med & max & cert.\\
\midrule
\multicolumn{7}{@{}l}{\textit{EuRoC \texttt{MH\_01}}, 4 windows}\\
Ours (SDP) & random & $0.88$ & $\mathbf{7.62}$ & $\mathbf{0.018}$ & $\mathbf{0.021}$ & $\mathbf{4/4}$\\
LM~\cite{forster} & truth & $0.88$ & $7.73$ & $0.019$ & $0.023$ & ---\\
LM~\cite{forster} & random & $4.31$ & $179.28$ & $0.019$ & $0.247$ & ---\\
\addlinespace
\multicolumn{7}{@{}l}{\textit{UrbanNav \texttt{medium urban}}, 2 windows}\\
Ours (SDP) & random & $29.06$ & $57.77$ & $\mathbf{0.0024}$ & $\mathbf{0.0034}$ & $\mathbf{2/2}$\\
LM~\cite{forster} & truth & $20.51$ & $40.67$ & $0.0030$ & $0.0043$ & ---\\
LM~\cite{forster} & random & $\mathbf{20.48}$ & $\mathbf{40.12}$ & $0.0030$ & $0.0043$ & ---\\
\addlinespace
\multicolumn{7}{@{}l}{\textit{UrbanNav \texttt{deep urban}}, 2 windows}\\
Ours (SDP) & random & $1.04$ & $1.40$ & $\mathbf{0.0035}$ & $\mathbf{0.0040}$ & $\mathbf{2/2}$\\
LM~\cite{forster} & truth & $1.04$ & $1.40$ & $0.0046$ & $0.0054$ & ---\\
LM~\cite{forster} & random & $1.04$ & $1.40$ & $0.0046$ & $0.0054$ & ---\\
\bottomrule
\end{tabular}
\end{table}

\subsection{Effect of Isotropic Modeling on Accuracy}
\label{sec:cmp}

To address the third question, we evaluate the $K=16$, $M=2$
configurations ($n=652$) in Table~\ref{tab:real}. We compare our estimator
with a standard exponential-map pre-integration factor~\cite{forster},
optimized using the Levenberg--Marquardt method in GTSAM. Because LM requires
an initial estimate, we test the two initialization strategies listed in
Table~\ref{tab:cmp}. The \emph{truth} setting initializes the full state near
the ground truth. For the \emph{random} setting, we run LM $200$ times from
randomly sampled initial states.

The proposed estimator gives accuracy comparable to LM on EuRoC and
\texttt{deep urban}. On \texttt{medium urban}, LM gives lower orientation
error, while the proposed estimator gives slightly lower position error.
Overall, the isotropic formulation retains comparable accuracy on the tested
windows while allowing the problem to be written in a certifiable form.

One randomly initialized LM run on a EuRoC window also illustrates the
sensitivity of local optimization to poor initialization, as shown in
Fig.~\ref{fig:teaser}. With the identity initialization used in this run, LM
converges to a gravity-flipped solution with a mean orientation error of
$178^\circ$. On the same window, the proposed method returns a mean orientation
error of $0.86^\circ$.

\section{Conclusion}
\label{sec:conc}

In this work, we proposed a QCQP-representable IMU pre-integration factor for
certifiable estimation. To obtain the required algebraic structure, we replaced
the exponential map with the Cayley map, derived polynomial residuals, and
introduced lifting variables to express the factor using a quadratic cost and
constraints. We also derived redundant constraints for these liftings to
tighten the SDP relaxation.
We validated the proposed formulation in two ways. First, synthetic ablation
studies confirmed that the derived redundant constraints improve SDP
relaxation exactness under different configurations. Second, we applied the
factor to certifiable GNSS--IMU smoothing on real-world data. The resulting
relaxations were tight, and the estimation problems were solved with verified
global optimality.
To the best of our knowledge, this is the first work to directly incorporate
IMU pre-integration into certifiable estimation. This extends the scope of
certifiable estimation to inertial sensing, an important modality in robotic
state estimation. The main remaining limitation is scalability, which is currently limited by
the cost of solving the SDP. Future work will explore how the problem
structure can be used to develop more scalable solvers.

\balance
\bibliographystyle{IEEEtran}
\bibliography{refs}

\end{document}